\documentclass[conference]{IEEEtran}
\IEEEoverridecommandlockouts
\usepackage[noadjust]{cite}
\usepackage{amsmath,amssymb,amsfonts}
\usepackage{algorithmic}
\usepackage{graphicx}
\usepackage{textcomp}
\usepackage{xcolor}

\usepackage{algorithm}
\usepackage{array}
\usepackage[caption=false]{subfig}

\usepackage{algorithmic}
\usepackage{float}
\usepackage{multirow}
\usepackage{booktabs} % for better table lines
\usepackage{tabularx}
\usepackage{multirow}
\usepackage{threeparttable}
\usepackage{bbding}
\usepackage{pifont}
\usepackage{wasysym}
\usepackage{amssymb}
\usepackage{makecell}
\usepackage{arydshln}
\usepackage[inline]{enumitem}
\usepackage[percent]{overpic}
\usepackage{xcolor}
\usepackage{amsmath, amssymb, amsfonts} % Dùng cho công thức toán
\usepackage{textcomp}

\def\BibTeX{{\rm B\kern-.05em{\sc i\kern-.025em b}\kern-.08em
    T\kern-.1667em\lower.7ex\hbox{E}\kern-.125emX}}
\begin{document}

\title{End-to-End Joint Training of Super-Resolution and Compression Artifact Removal for Industrial Machine Vision
}

\author{
\IEEEauthorblockN{Dang BuiDinh\textsuperscript{1}, Sang NguyenQuang\textsuperscript{2}, Anh NguyenPhuong\textsuperscript{1}, Wen-Hsiao Peng\textsuperscript{2}, and Xiem HoangVan\textsuperscript{1*}
% \thanks{This research is funded by Vietnam National Foundation for Science and Technology Development (NAFOSTED) under grant number NCUD.02-2024.09.}
}
\IEEEauthorblockA{\textsuperscript{1}\textit{Faculty of Electronics and Telecommunications, VNU University of Engineering and Technology, Hanoi, Vietnam}}
\IEEEauthorblockA{\textsuperscript{2}\textit{Department of Computer Science, National Yang Ming Chiao Tung University, Hsinchu, Taiwan}}
\IEEEauthorblockA{\textsuperscript{*}\textit{Corresponding author: xiemhoang@vnu.edu.vn}}
}
\maketitle

\begin{abstract}
% In industrial machine vision, images captured at the edge are often downsampled and compressed to reduce bandwidth, introducing resolution loss and compression artifacts that severely degrade downstream tasks such as defect detection. Existing quality enhancement methods typically apply super-resolution (SR) or compression artifact removal (CAR) independently or in a sequential manner and are largely optimized for human visual fidelity rather than machine perception. This paper proposes an end-to-end quality enhancement framework that jointly integrates SR and CAR with a unified training strategy that optimizes both human-perception and machine-perception objectives. We investigate fixed and uncertainty-based dynamic loss weighting to balance reconstruction quality and task performance. Extensive experiments on PCB Defect, DeepPCB, and NEU-DET datasets show that the proposed jointly trained model consistently outperforms standalone and separately trained SR–CAR pipelines. In particular, under severe JPEG compression, our method improves defect detection performance from 0.777 to 0.866 mAP@50 over sequential SR–CAR baselines, with only a slight reduction in PSNR. These results demonstrate that joint optimization is critical for compression-aware industrial machine vision systems.

In industrial machine vision, images captured by edge devices are often downsampled and compressed to reduce bandwidth, introducing resolution loss and compression artifacts that severely degrade the performance of downstream tasks, such as defect detection. Existing quality enhancement methods typically apply super-resolution (SR) or compression artifact removal (CAR) independently or in a sequential manner and are largely optimized for human visual fidelity rather than machine perception. This paper proposes an end-to-end quality enhancement framework that integrates SR and CAR with a unified training strategy that optimizes both human-perception and machine-perception objectives. We investigate fixed-weight and uncertainty-based dynamic-weight loss functions to balance reconstruction quality and machine vision task performance. Extensive experiments on PCB Defect, DeepPCB, and NEU-DET datasets show that the proposed jointly trained model generally outperforms standalone and separately trained SR–CAR pipelines. In particular, under severe JPEG compression, our method improves defect detection performance from 0.461 to 0.887 mAP@50 on PCB Defect over sequential SR–CAR baselines, with only a slight reduction in PSNR. These results demonstrate that joint optimization is critical for compression-aware industrial machine vision systems.
\end{abstract}


\section{Introduction}
\label{sec:introduction}

% Machine vision plays a crucial role in many industrial applications, such as automated inspection, defect detection, and intelligent manufacturing. Recent advances in deep learning have significantly improved the accuracy and robustness of these systems. However, deploying computationally intensive models on edge devices remains challenging due to limited computational and memory resources. Although model inference can be offloaded to server-side systems, transmitting raw high-resolution images requires substantial bandwidth, which is often impractical for real-time applications.

Machine vision is essential in industrial applications such as automated inspection and defect detection. Despite recent advances in deep learning, deploying complex models on edge devices remains challenging due to resource constraints. Although model inference can be offloaded to server-side systems, transmitting raw high-resolution images requires substantial bandwidth, which is often impractical for real-time applications.

To reduce transmission bandwidth, captured images are typically downsampled and compressed using lossy compression schemes~\cite{wallace1992jpeg, vvc} before transmission. While effective for data reduction, this process inevitably causes resolution loss and compression artifacts, resulting in severe image degradation. Such degradation substantially affects the performance of downstream machine vision tasks, especially in fine-grained defect detection.

Several recent learning-based studies~\cite{esrpcb, e2dsr,nafnet, cajnn,arcnn, srcnn, edsr, swin2sr, mfqe_vvc, ovqe_vvc, surveyvqe_atc2025, mdfi} have investigated the use of super-resolution (SR) and compression artifact removal (CAR) to mitigate compression-induced degradation. 
However, these approaches still exhibit notable limitations. First, most existing methods apply SR or CAR independently~\cite{esrpcb, nafnet, arcnn}, or combine both components while training them separately. 
Second, most existing methods mainly focus on optimizing human visual perception by minimizing pixel-level reconstruction errors~\cite{cajnn,  reca}. 
While effective for visual quality, such objectives are not necessarily optimal for machine perception, which is more sensitive to feature-level distortions.
\begin{figure}[!t]
    \centering
    \includegraphics[width=\linewidth]{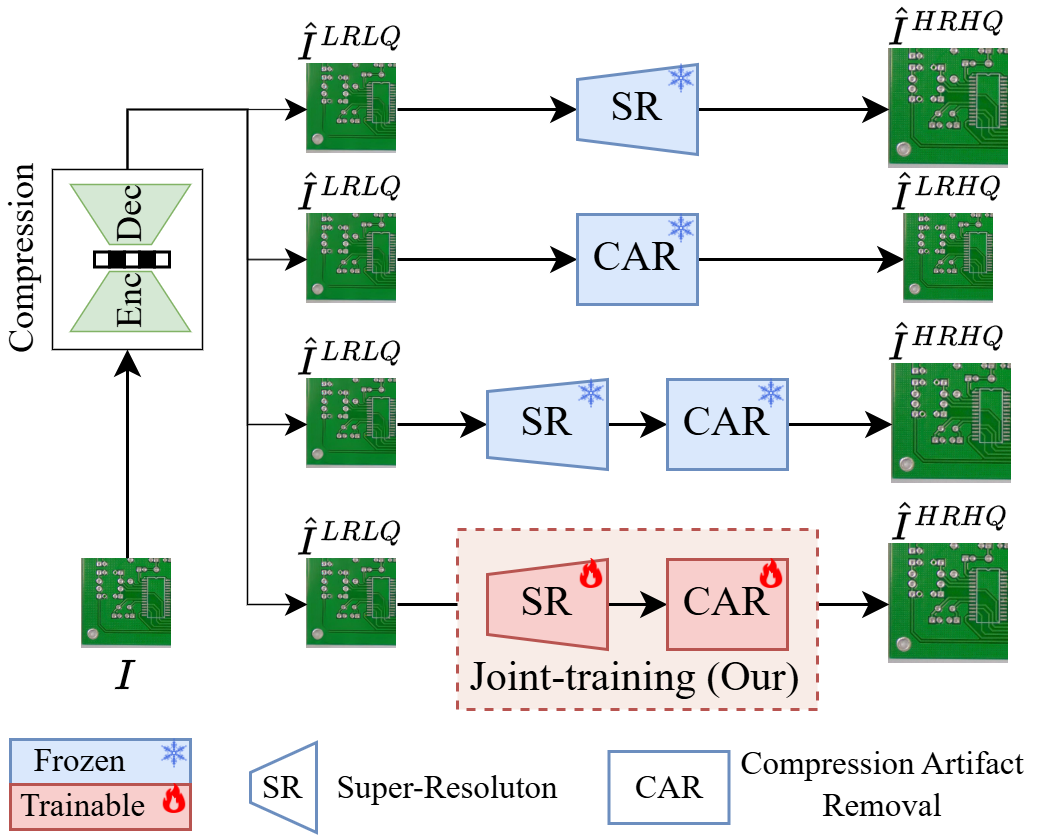}
    \caption{Comparison between our proposed joint-training
method and separate training approaches for SR and CAR.}
    % \vspace{-1em}
    \label{fig:combined}
\vspace{-0.4cm}
\end{figure}

Motivated by these limitations, we propose a novel quality enhancement (QE) framework designed to mitigate compression degradation for machine vision tasks. 
Our approach introduces an end-to-end joint model that integrates SR and CAR with multiple architectural variants, as illustrated in Fig.~\ref{fig:combined}. 
Furthermore, we develop a unified training pipeline with integrated loss functions that simultaneously optimize both human and machine-perception objectives.

In summary, our main contributions are as follows:
\begin{itemize}
    \item We propose an end-to-end QE framework that jointly integrates SR and CAR in two processing orders.
    \item We introduce a joint optimization strategy that combines human-perception and machine-perception losses.
    \item Extensive experiments on defect detection tasks under various compression settings validate the effectiveness and robustness of the proposed framework.
\end{itemize}

% The remainder of this paper is organized as follows. Section~\ref{sec:related} reviews the related works. Section~\ref{sec:proposed_method} describes the proposed QE framework with joint training strategy in detail. Section~\ref{sec:evaluation} presents the experimental settings and results. Finally, Section~\ref{sec:conclusions} concludes the paper and outlines future research directions.

\section{Related Work}
\label{sec:related}
\textbf{Super-Resolution:} Super-resolution (SR) aims to reconstruct high-resolution images from low-resolution inputs. Early interpolation methods, such as bilinear interpolation, often produced overly smooth results lacking fine details. Deep learning has significantly advanced the field, beginning with SRCNN~\cite{srcnn}, which demonstrated the effectiveness of convolutional neural networks (CNNs). Deeper models, including EDSR~\cite{edsr} and ESRPCB~\cite{esrpcb}, further improved reconstruction quality. GAN-based methods~\cite{real_esrgan} improve perceptual realism, whereas transformer-based models~\cite{swin2sr} effectively capture long-range dependencies. However, these approaches often incur higher computational costs, which may limit their deployment on resource-constrained devices.

\textbf{Compression Artifact Removal:} Early restoration methods relied on handcrafted priors, such as sparse coding~\cite{dong2011sparsity}, but often struggled with complex textures. AR-CNN~\cite{arcnn} subsequently introduced deep learning for compression artifact reduction. Encoder--decoder architectures, particularly U-Net-like architecture~\cite{nafnet,swin2sr,retinexformer}, later became prominent due to their efficient hierarchical multi-scale processing. By capturing both local details and global structures, they are widely used in CNN-based and transformer-based CAR models.

\section{Proposed Method}
\label{sec:proposed_method}
\subsection{Preliminary}
\label{ssec:Preliminary}

We consider an industrial image quality enhancement framework, as illustrated in Fig.~\ref{fig:combined}. To transmit or store a captured image $I$, encoders typically downsample and compress it (e.g., JPEG~\cite{wallace1992jpeg} and VVC~\cite{vvc}). Because of downsampling and lossy compression, the decoded image has reduced spatial resolution and noticeable compression artifacts.
To mitigate these degradations and facilitate downstream machine vision tasks (e.g., defect detection), we employ QE techniques such as SR and CAR. We evaluate the impact of QE methods under five configurations: (1) directly using the low-quality reconstructed image $I^{LRLQ}$; (2) applying SR only; (3) applying CAR only; (4) sequentially applying CAR followed by SR; and (5) applying SR followed by CAR.
Experiments are conducted on the PCB Defect dataset~\cite{pcbdataset}. For the QE modules, we use ESRPCB~\cite{esrpcb} for SR and a transformer-based UNet backbone~\cite{retinexformer} for CAR, training each module separately using a pixel-wise reconstruction loss

Table~\ref{tab:pre_map} reports the defect detection performance in terms of mean Average Precision at an Intersection over Union threshold of 0.5 (mAP@50). 
% Without QE (i.e., Setting (1)), the detection performance is severely degraded. Applying either SR (i.e., Setting (2)) or CAR (i.e., Setting (3)) yields substantial improvements. However, directly cascading SR and CAR (i.e., Settings (4) and (5)) does not further improve performance and may even degrade it.
While SR or CAR (i.e., Settings (2) and (3)) alone improves detection performance, directly cascading them (i.e., Settings (4) and (5)) provides no further gain and may even degrade performance.

% Based on these findings, we propose an end-to-end quality enhancement framework that integrates SR and CAR to recover high-quality images while suppressing compression artifacts. In addition, to better serve industrial machine vision applications, we introduce a joint optimization strategy that explicitly considers both human and machine perception objectives. To the best of our knowledge, this is the first unified training strategy to address these objectives within a combined SR--CAR framework, and Table~\ref{tab:comparison} summarizes its key distinctions from representative related methods. Implementation details are described in Section~\ref{ssec:detail_implement}.

Based on these findings, we propose an end-to-end QE framework that jointly integrates SR and CAR and optimizes both human- and machine-perception objectives. To the best of our knowledge, this is the first unified training strategy for a combined SR--CAR framework. Table~\ref{tab:comparison} highlights its key differences from related methods, and implementation details are provided in Section~\ref{ssec:detail_implement}.

\begin{table}[t]
\centering
\caption{Defect detection performance of different QE strategies on the PCB Defect dataset.}
\label{tab:pre_map}
\footnotesize
\resizebox{\linewidth}{!}
{%
\begin{tabular}{ccccccc} 
\toprule
Setting & QE Strategy    & QF20  & QF40  & QF60  & QF80  & Average  \\ 
\midrule
(1)      & w/o QE & 0.052 & 0.151 & 0.250 & 0.436 & 0.222    \\
(2)      & SR only & 0.214 & 0.476 & 0.618 & 0.751 & 0.515    \\
(3)      & CAR only  & 0.155 & 0.330 & 0.452 & 0.612 & 0.387    \\
(4)      & CAR + SR         & 0.124 & 0.318 & 0.435 & 0.615 & 0.373    \\
(5)      & SR + CAR         & 0.204 & 0.461 & 0.606 & 0.760 & 0.508    \\
\bottomrule
\end{tabular}%
}
\scriptsize 
\begin{flushleft}
CAR + SR: CAR is performed before SR; SR + CAR: SR is performed before CAR.
\end{flushleft}
% \vspace{-0.4cm}
\end{table}

\begin{table}[!t]
\centering
% \fontsize{7.8pt}{7.8pt}\selectfont
\caption{Comparison of the proposed framework with representative related methods.}
\label{tab:comparison}
% \color{red}
% \setlength{\tabcolsep}{1.3pt}
\resizebox{\linewidth}{!}{%
\begin{tabular}{lccccccc}
\toprule
Method
& \cite{esrpcb}
& \cite{reca}
% & \cite{arsr}
& \cite{nafnet}
& \cite{cajnn}
& \cite{uw}
& Our \\ \midrule
Joint combination training   &              & \checkmark   & \checkmark   & \checkmark   &               & \checkmark   \\
Human perception loss        & \checkmark   & \checkmark   & \checkmark   & \checkmark   &               & \checkmark   \\
Machine perception loss      &              &              &              &              & \checkmark    & \checkmark   \\
Downstream vision task       & \checkmark   &              &              &              & \checkmark    & \checkmark   \\
\bottomrule
\end{tabular}%
}
\vspace{-0.4cm}
\end{table}
\begin{figure*}[!t]
    \centering
    \includegraphics[width=0.9\linewidth]{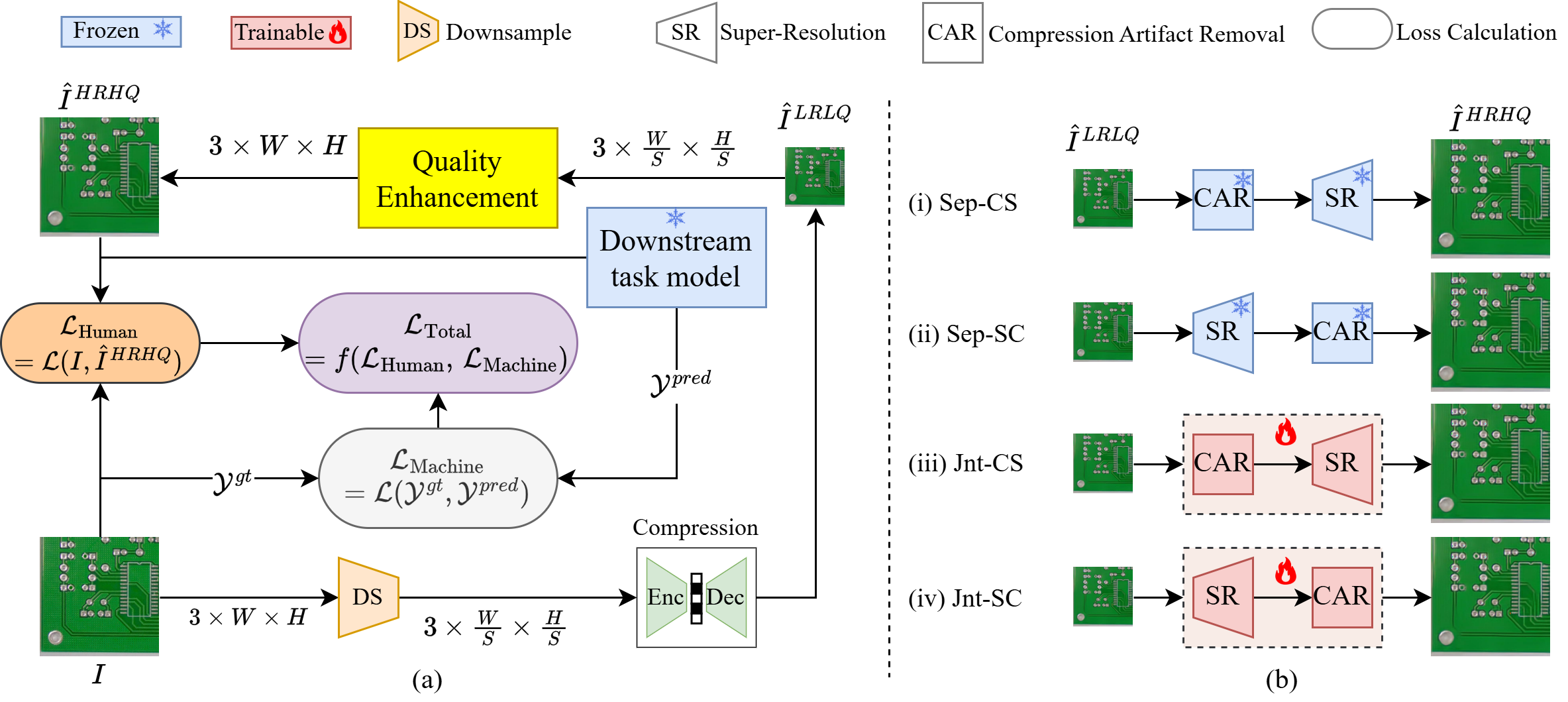}
    \caption{
    %Overall architecture of our proposed training strategy across various detailed quality enhancement configurations.
    Overall architecture of the proposed QE framework.
    (a) End-to-end joint training pipeline, where the QE module is optimized under both human-perception and machine-perception losses.
    (b) Different QE configurations, including separately trained models (Sep-CS and Sep-SC) and jointly trained models (Jnt-CS and Jnt-SC).
    }
    % \vspace{-1em}
    \label{fig:overall}
\vspace{-0.4cm}
\end{figure*}

\subsection{System Overview}
% \newpage
% Fig.~\ref{fig:overall}~(a) illustrates the proposed framework with our novel training strategy. The image captured by camera $I \in \mathbb{R}^{3 \times W \times H}$ with three color channels and spatial dimensions of width $W$ and height $H$ is free of compression artifacts and high-resolution. The corresponding ground-truth label $\mathcal{Y}^{gt}$ is manually annotated. To save transmission bandwidth and storage, the raw image $I$ is first downsampled by a scale factor of $S$, and compressed into a bitstream using a lossy image compression approach (e.g., JPEG). At the decoder side, the decoding process reconstructs a low-resolution, degraded-quality image $\hat{I}^{LRLQ} \in \mathbb{R}^{3 \times \frac{W}{S} \times \frac{H}{S}}$. To enhance the quality of the reconstructed image, a QE model is employed. The design of the QE model is depicted in Fig.~\ref{fig:overall}~(b), with implementation details presented in Section.~\ref{ssec:detail_implement}. The QE model output is $\hat{I}^{HRHQ} \in \mathbb{R}^{3 \times W \times H}$, which exhibits high resolution and minimal compression artifacts. The downstream task (e.g. defect detection and classification) is then performed on this enhanced image, yielding the predicted label $\mathcal{Y}^{pred}$.
% \newpage
Fig.~\ref{fig:overall}(a) illustrates the proposed framework with a novel training strategy. The captured image $I \in \mathbb{R}^{3 \times W \times H}$ is high-resolution and free of compression artifacts, while its corresponding ground-truth label $\mathcal{Y}^{gt}$ is manually annotated. To reduce transmission bandwidth and storage, the raw image $I$ is downsampled by a scale factor of $S$ and compressed using a lossy codec (e.g., JPEG). At the decoder, it is reconstructed as a low-resolution, degraded-quality image $\hat{I}^{LRLQ} \in \mathbb{R}^{3 \times \frac{W}{S} \times \frac{H}{S}}$. The QE model, shown in Fig.~\ref{fig:overall}(b) and detailed in Section~\ref{ssec:detail_implement}, produces $\hat{I}^{HRHQ} \in \mathbb{R}^{3 \times W \times H}$ with restored resolution and reduced artifacts. The enhanced image is then used for downstream tasks (e.g., defect detection and classification), yielding $\mathcal{Y}^{pred}$.

In optimizing QE models, Mean Squared Error (MSE) or Charbonnier Loss~\cite{Charbonnier_Loss} is typically employed to maximize the fidelity of reconstructed images. However, this creates a fundamental trade-off: models trained with a human-oriented loss ($\mathcal{L}_{\text{Human}}$) frequently underperform on downstream machine vision tasks. Conversely, utilizing a machine-oriented loss ($\mathcal{L}_{\text{Machine}}$) improves automated analysis but generates unnatural artifacts that severely degrade perceptual quality.

In this paper, our QE models are trained with a novel training strategy that jointly optimizes reconstruction quality and machine task performance through a combined objective function $\mathcal{L}_{\text{Joint}}$, defined as:
\begin{equation}
    \mathcal{L}_{\text{Joint}} = f\!\left(\mathcal{L}_{\text{Human}},\, \mathcal{L}_{\text{Machine}}\right),
\end{equation}
where $f(\cdot)$ denotes a generic loss aggregation function for the human-perception and machine-perception losses.

\subsection{Implementation Details}
\label{ssec:detail_implement}

% To optimize human visual quality, the QE network is trained to minimize the distortion between the reconstructed image $\hat{I}^{HRHQ}$ and the raw image $I$. We define $\mathcal{L}_{\text{Human}}$ as the objective function for this training strategy and adopt Charbonnier loss~\cite{Charbonnier_Loss} in our implementation:

To optimize visual quality, we define $\mathcal{L}_{\text{Human}}$ using the Charbonnier loss~\cite{Charbonnier_Loss} between the reconstructed image $\hat{I}^{HRHQ}$ and the original image $I$:
\begin{equation}
    \mathcal{L}_{\text{Human}}
=
\frac{1}{N}
\sum_{i=1}^{N}
\sqrt{
\left(I_i-\hat{I}^{HRHQ}_i\right)^2+\epsilon^2
},
\end{equation}
where $N$ is the number of image elements, and $\epsilon$ is a small constant, typically set to $10^{-8}$, ensuring numerical stability near zero.

On the other hand, to maximize machine perception performance, the model is trained with a machine-oriented objective function, denoted as $\mathcal{L}_{\text{Machine}}$. In our design, with the defect detection task, $\mathcal{L}_{\text{Machine}}$ is a summation of three components: localization loss $\mathcal{L}_{\text{box}}$, classification loss $\mathcal{L}_{\text{cls}}$, and distribution focal loss $\mathcal{L}_{\text{dfl}}$, which can be expressed as:
\begin{equation}
    \mathcal{L}_{\text{Machine}}=  \mathcal{L}_{\text{cls}} + \mathcal{L}_{\text{box}} + \mathcal{L}_{\text{dfl}}.
\end{equation}
During joint training, the pretrained YOLOv8 detector is frozen, and only the QE model is updated through $\mathcal{L}_{\text{Machine}}$.
% As previously discussed in Section~\ref{ssec:Preliminary}, simply combining SR and CAR without a proper training strategy for downstream machine vision tasks not only fails to improve performance but also reduces model efficiency. To address this, we propose a joint training strategy to optimize the QE model that combines SR and CAR networks in an end-to-end training pipeline with two integrated objective functions: (1) Fixed-Weight Loss Function and (2) Dynamic-Weight Loss Function. Both approaches are designed to effectively balance human visual quality with machine perception performance.

As discussed in Section~\ref{ssec:Preliminary}, simply combining SR and CAR without task-oriented training may degrade both performance and efficiency. To address this, we propose an end-to-end joint training strategy with two objectives: (1) Fixed-Weight Loss Function and (2) Dynamic-Weight Loss Function, both balancing human visual quality and machine-perception performance.

\begin{table}[!t]
\centering
\footnotesize
\caption{QE model configurations and training strategies.}
\label{tab:model_variant}
% \resizebox{\linewidth}{!}{%
\begin{tabular}{cccc} 
\toprule
Model                        & Description        & Training      \\ 
\midrule
SR only                   & Only perform SR    & Separate  \\
CAR only                                                                      & Only perform CAR   & Separate  \\
\midrule
Sep-CS                                                & Perform CAR first, followed by SR & Separate  \\
Sep-SC                    & Perform SR first, followed by CAR & Separate  \\
\midrule
Jnt-CS                                                          & Perform CAR first, followed by SR & Joint     \\
Jnt-SC                        & Perform SR first, followed by CAR & Joint     \\

\bottomrule
\end{tabular}
% }
\vspace{-0.2cm}
\end{table}
\begin{table*}[!t]
\centering
% \caption{Comparison of mAP@50 and F1-scores across different datasets and QF values under various loss types and QE model configurations. Notation: mAP@50/F1-scores.}
\caption{Defect detection performance across datasets, JPEG quality factors, QE configurations, and loss settings. Results are reported as mAP@50/F1 score.}
\label{tab:map50}
\resizebox{\linewidth}{!}{%
\renewcommand{\arraystretch}{0.9}
\begin{tabular}{cclccccccc} 
\toprule
Dataset                                        & QF                  & Loss Type                    & w/o QE                          & SR only                                          & CAR only      & Sep-CS        & Sep-SC        & \makecell[c]{Jnt-CS}                       & \makecell{Jnt-SC}                             \\ 
\midrule
\multirow{12}{*}{PCB Defect~\cite{pcbdataset}} & \multirow{3}{*}{20} & $\mathcal{L}_\text{Human}$   & \multirow{3}{*}{0.052 / 0.081}  & 0.214 / 0.266                                    & 0.155 / 0.191 & 0.124 / 0.158 & 0.204 / 0.259 & 0.169 / 0.214                                     & 0.225 / 0.275                                      \\
                                               &                     & $\mathcal{L}_\text{Fixed}$   &                                 & 0.611 / 0.635                                    & 0.579 / 0.195 & 0.475 / 0.418 & 0.471 / 0.344 & 0.640 / 0.663                                     & 0.634 / 0.644                                      \\
                                               &                     & $\mathcal{L}_\text{Dynamic}$ &                                 & 0.608 / 0.617                                    & 0.573 / 0.611 & 0.512 / 0.473 & 0.446 / 0.373 & \textcolor{blue}{0.661 }/ \textcolor{red}{0.696}  & \textcolor{red}{0.669 }/ \textcolor{blue}{0.667}   \\ 
\cmidrule(lr){2-10}
                                               & \multirow{3}{*}{40} & $\mathcal{L}_\text{Human}$   & \multirow{3}{*}{0.151 / 0.192}  & 0.476 / 0.529                                    & 0.330 / 0.375 & 0.318 / 0.369 & 0.461 / 0.513 & 0.395 / 0.457                                     & 0.440 / 0.501                                      \\
                                               &                     & $\mathcal{L}_\text{Fixed}$   &                                 & 0.861 / 0.856                                    & 0.801 / 0.813 & 0.710 / 0.610 & 0.764 / 0.637 & 0.872 / 0.870                                     & \textcolor{blue}{0.877 }/ 0.880                    \\
                                               &                     & $\mathcal{L}_\text{Dynamic}$ &                                 & 0.866 / 0.863                                    & 0.804 / 0.808 & 0.741 / 0.662 & 0.814 / 0.718 & 0.876 / \textcolor{blue}{0.876}                   & \textcolor{red}{0.887 }/ \textcolor{red}{0.878}    \\ 
\cmidrule(lr){2-10}
                                               & \multirow{3}{*}{60} & $\mathcal{L}_\text{Human}$   & \multirow{3}{*}{0.250 / 0.287}  & 0.618 / 0.663                                    & 0.452 / 0.497 & 0.435 / 0.497 & 0.606 / 0.653 & 0.544 / 0.601                                     & 0.591 / 0.646                                      \\
                                               &                     & $\mathcal{L}_\text{Fixed}$   &                                 & 0.915 / 0.910                                    & 0.880 / 0.888 & 0.800 / 0.714 & 0.850 / 0.763 & 0.927 / 0.928                                     & 0.933 / 0.928                                      \\
                                               &                     & $\mathcal{L}_\text{Dynamic}$ &                                 & 0.924 / 0.916                                    & 0.879 / 0.888 & 0.830 / 0.720 & 0.905 / 0.838 & \textcolor{blue}{0.938 }/ \textcolor{blue}{0.934} & \textcolor{red}{0.943 }/ \textcolor{red}{0.939}    \\ 
\cmidrule(lr){2-10}
                                               & \multirow{3}{*}{80} & $\mathcal{L}_\text{Human}$   & \multirow{3}{*}{0.436 / 0.479~} & 0.751 / 0.771                                    & 0.612 / 0.647 & 0.615 / 0.659 & 0.760 / 0.779 & 0.752 / 0.777                                     & 0.760 / 0.791                                      \\
                                               &                     & $\mathcal{L}_\text{Fixed}$   &                                 & \textcolor{blue}{0.959 }/ 0.961                  & 0.928 / 0.939 & 0.872 / 0.788 & 0.940 / 0.880 & \textcolor{blue}{0.959 }/ 0.961                   & \textcolor{red}{0.965 }/ \textcolor{red}{0.970}    \\
                                               &                     & $\mathcal{L}_\text{Dynamic}$ &                                 & 0.957 / \textcolor{blue}{0.963}                  & 0.924 / 0.937 & 0.898 / 0.835 & 0.941 / 0.913 & \textcolor{blue}{0.959 }/ \textcolor{blue}{0.963} & \textcolor{red}{0.965 }/ \textcolor{red}{0.970}    \\ 
\midrule
\multirow{6}{*}{DeepPCB~\cite{deeppcb}}        & \multirow{3}{*}{20} & $\mathcal{L}_\text{Human}$   & \multirow{3}{*}{0.363 /~0.448}  & 0.857 / 0.897                                    & 0.555 / 0.650 & 0.483 / 0.603 & 0.818 / 0.866 & 0.870 / 0.910                                     & 0.845 / 0.887                                      \\
                                               &                     & $\mathcal{L}_\text{Fixed}$   &                                 & 0.931 / 0.937                                    & 0.892 / 0.905 & 0.884 / 0.906 & 0.918 / 0.920 & 0.935 / 0.942                                     & 0.943 / 0.946                                      \\
                                               &                     & $\mathcal{L}_\text{Dynamic}$ &                                 & \textcolor{blue}{0.946 }/ \textcolor{red}{0.953} & 0.913 / 0.922 & 0.581 / 0.634 & 0.910 / 0.926 & 0.940 / 0.941                                     & \textcolor{red}{0.948 }/ \textcolor{blue}{0.948}   \\ 
\cmidrule(lr){2-10}
                                               & \multirow{3}{*}{40} & $\mathcal{L}_\text{Human}$   & \multirow{3}{*}{0.477 / 0.565}  & 0.891 / 0.913                                    & 0.602 / 0.677 & 0.455 / 0.593 & 0.887 / 0.919 & 0.892 / 0.915                                     & 0.889 / 0.915                                      \\
                                               &                     & $\mathcal{L}_\text{Fixed}$   &                                 & 0.953 / 0.955                                    & 0.905 / 0.918 & 0.916 / 0.924 & 0.915 / 0.917 & 0.947 / 0.952                                     & \textcolor{blue}{0.956 }/ 0.953                    \\
                                               &                     & $\mathcal{L}_\text{Dynamic}$ &                                 & \textcolor{red}{0.958 }/ \textcolor{red}{0.962}  & 0.934 / 0.936 & 0.396 / 0.472 & 0.935 / 0.947 & 0.948 / 0.952                                     & \textcolor{red}{0.958 }/ \textcolor{blue}{0.959}   \\ 
\midrule
\multirow{6}{*}{NEU-DET~\cite{neudet}}         & \multirow{3}{*}{20} & $\mathcal{L}_\text{Human}$   & \multirow{3}{*}{0.212 / 0.283}  & 0.377 / 0.447                                    & 0.365 / 0.438 & 0.332 / 0.401 & 0.345 / 0.409 & 0.387 / 0.458                                     & 0.411 / 0.489                                      \\
                                               &                     & $\mathcal{L}_\text{Fixed}$   &                                 & 0.379 / 0.480                                    & 0.386 / 0.496 & 0.331 / 0.418 & 0.311 / 0.400 & 0.331 / 0.418                                     & \textcolor{blue}{0.420} / \textcolor{blue}{0.527}  \\
                                               &                     & $\mathcal{L}_\text{Dynamic}$ &                                 & 0.387 / 0.490                                    & 0.378 / 0.490 & 0.349 / 0.468 & 0.297 / 0.392 & 0.366 / 0.480                                     & \textcolor{red}{0.429 }/ \textcolor{red}{0.535}    \\ 
\cmidrule(lr){2-10}
                                               & \multirow{3}{*}{40} & $\mathcal{L}_\text{Human}$   & \multirow{3}{*}{0.303 / 0.366}  & 0.434 / 0.511                                    & 0.401 / 0.480 & 0.419 / 0.501 & 0.402 / 0.473 & 0.451 / 0.519                                     & 0.463 / 0.530                                      \\
                                               &                     & $\mathcal{L}_\text{Fixed}$   &                                 & 0.468 / 0.554                                    & 0.467 / 0.562 & 0.441 / 0.521 & 0.446 / 0.529 & 0.477 / 0.568                                     & \textcolor{red}{0.519~}/ \textcolor{red}{0.595}    \\
                                               &                     & $\mathcal{L}_\text{Dynamic}$ &                                 & 0.460 / 0.549                                    & 0.449 / 0.560 & 0.427 / 0.540 & 0.405 / 0.492 & 0.482 / 0.565                                     & \textcolor{blue}{0.491 }/ \textcolor{blue}{0.593}  \\
\bottomrule
\end{tabular}%
}
% \scriptsize 
% \begin{flushleft}
% The \textcolor{red}{Red} and \textcolor{blue}{Blue} denote the best and second best results.
% \end{flushleft}
\vspace{-0.4cm}
\end{table*}
\subsubsection{Fixed-Weight Loss Function}

We design the loss function for our joint training strategy as a linear combination of $\mathcal{L}_{\text{Human}}$ and $\mathcal{L}_{\text{Machine}}$ with fixed weights. Specifically, the loss function is formulated as:
\begin{equation}
    \mathcal{L}_{\text{Joint}} = \mathcal{L}_{\text{Fixed}} = \mathcal{L}_{\text{Human}} \times w_H + \mathcal{L}_{\text{Machine}} \times w_M,
\end{equation}
where $w_H$ and $w_M$ denote the weighting coefficients for the human-perception loss ($\mathcal{L}_{\text{Human}}$) and the machine-perception loss ($\mathcal{L}_{\text{Machine}}$), respectively, controlling their relative contributions during optimization.
During training, we observe that the magnitude of $\mathcal{L}_{\text{Machine}}$ is often more than 100 times larger than that of $\mathcal{L}_{\text{Human}}$. To mitigate this imbalance, we empirically set the loss weights $w_H = 1$ and $w_M = 0.01$, thereby effectively balancing the contribution of $\mathcal{L}_{\text{Human}}$ while still preserving task-driven optimization.

\subsubsection{Dynamic-Weight Loss Function}
Although fixed weighting provides a simple way to balance human- and machine-perception objectives, manually chosen weights may not generalize across datasets or training stages. We therefore adopt task-uncertainty-based Uncertainty Weighting (UW)~\cite{uw}, which automatically adjusts the contribution of each loss term during training. The dynamic loss is formulated as:
\begin{equation}
% \scriptsize
    \begin{aligned}
        \mathcal{L}_{\text{Joint}} &= \mathcal{L}_{\text{Dynamic}}\\& =\frac{1}{2\sigma_H^{2}} \mathcal{L}_{\text{Human}}
        + \frac{1}{2\sigma_M^{2}} \mathcal{L}_{\text{Machine}}
        + \log(\sigma_H \sigma_M),
    \end{aligned}
\end{equation}
where $\sigma_H$ and $\sigma_M$ are trainable uncertainty parameters jointly optimized with the network to adaptively balance human- and machine-perception objectives.
% where $\sigma_H$ and $\sigma_M$ are trainable parameters representing the uncertainty of the human- and machine-perception objectives, respectively. They are jointly optimized with the network parameters to adaptively balance the two objectives, eliminating manual weight tuning and yielding more robust convergence and improved generalization for machine vision tasks across varying compression conditions.

% where $\sigma_H$ and $\sigma_M$ are the trainable parameters and jointly optimized with the network parameters during end-to-end training.
% where $\sigma_H$ and $\sigma_M$ are trainable parameters representing the task-dependent uncertainty associated with the human-perception and machine-perception objectives, respectively. These parameters are jointly optimized with the network weights during training to adaptively balance perceptual and task-driven objectives based on relative uncertainty. This dynamic weighting eliminates the need for manual tuning, yielding more robust convergence and improved generalization for machine vision tasks across varying compression conditions.

\section{Experiments} \label{sec:evaluation}
\subsection{Experimental Settings}
% We follow the system architecture shown in Fig.~\ref{fig:overall}~(a) to conduct our experiments. The input image is first downsampled with a scale factor of $S=4$. JPEG compression~\cite{wallace1992jpeg} is then applied to simulate the compression process, and four quality factors (QFs) (i.e., 20, 40, 60, and 80) are used to generate reconstructed images with various degradation levels. The quality enhancement module is evaluated under six distinct configurations, as outlined in Table~\ref{tab:model_variant}. Specifically:
Following Fig.~\ref{fig:overall}~(a), each input image is downsampled by a factor of $S=4$ and compressed using JPEG~\cite{wallace1992jpeg}. Four quality factors (QFs) (i.e., 20, 40, 60, and 80) are used to produce different degradation levels. Table~\ref{tab:model_variant} summarizes the six evaluated QE configurations:
\begin{itemize}
    % \item \textbf{Single QE model:} The QE model simply consists of either the SR network or the CAR network alone. In the \textit{SR only} configuration, the output image is in high-resolution, whereas in the \textit{CAR only} setting, the output remains low-resolution but with improved visual quality.
    % \item \textbf{Combined QE model with separate training:} The QE model is a simple combination of SR and CAR. In this configuration, SR and CAR are two independently trained models. We evaluate two sequential orders: SR followed by CAR (\textit{Sep-SC}), and CAR followed by SR (\textit{Sep-CS}).
    % \item \textbf{Combined QE model with joint training:} In contrast to the independent training of SR and CAR models, the combined QE model is trained using our proposed joint training strategy. We evaluate the same two sequential orders: SR followed by CAR (\textit{Jnt-SC}), and CAR followed by SR (\textit{Jnt-CS}).

    \item \textbf{Single QE model:} Applies either SR or CAR independently. The \textit{SR only} configuration yields a high-resolution output, while \textit{CAR only} retains the original resolution but reduces artifacts.
    \item \textbf{Combined QE model with separate training:} Combines SR and CAR models, which are two independently trained models. We assess two orders: SR followed by CAR (\textit{Sep-SC}) and CAR followed by SR (\textit{Sep-CS}).
    \item \textbf{Combined QE model with joint training:} Employs our proposed joint training strategy for the combined model. We also evaluate both \textit{Jnt-SC} and \textit{Jnt-CS}.
\end{itemize}
\begin{figure}[!t]
    \centering
    \includegraphics[width=1\linewidth]{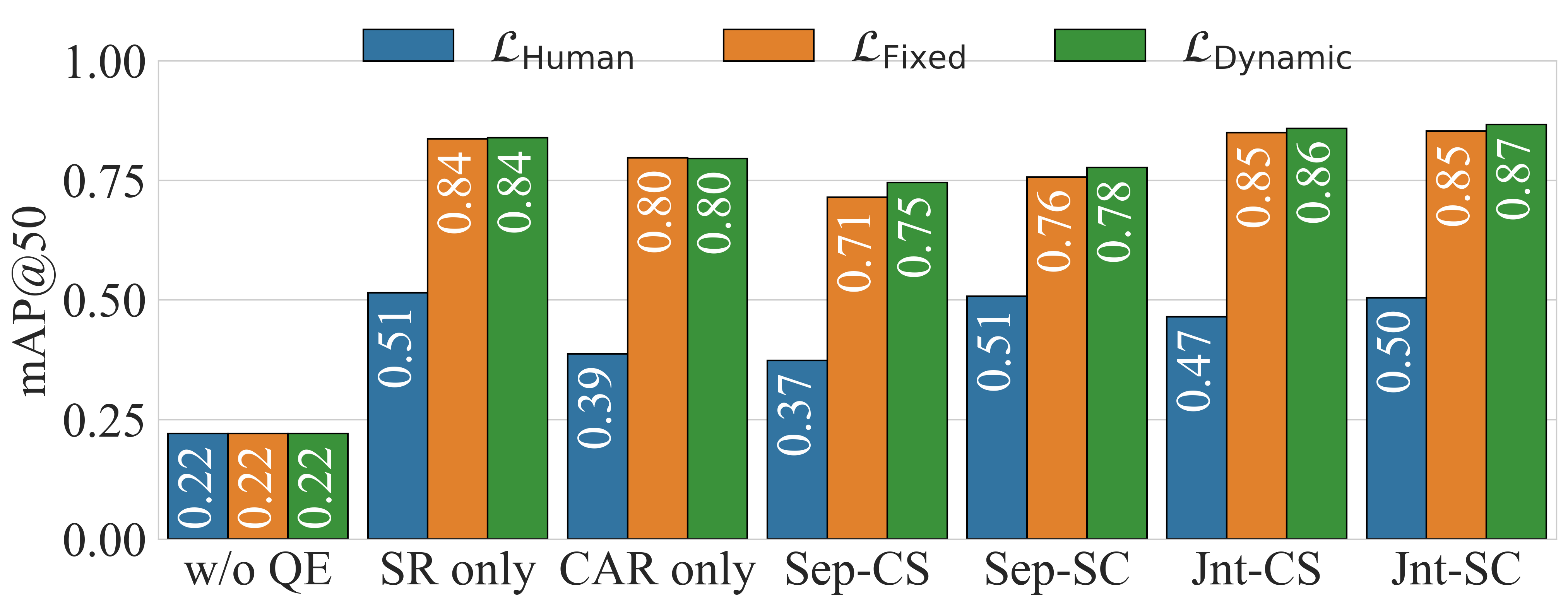}
    \caption{Average mAP@50 of different QE configurations on the PCB Defect dataset across all evaluated QFs.}    % \vspace{-1em}
    \label{fig:avg_map}
\vspace{-0.5cm}
\end{figure}
\setlength{\fboxsep}{1pt}

\begin{figure*}[!t]
    \centering
    % giảm khoảng cách giữa các cột
    \setlength{\tabcolsep}{1pt}
    % bỏ khoảng cách giữa các hàng
    \renewcommand{\arraystretch}{1}
    \begin{tabular}{cccccc}
         Ground Truth& w/o QE &  Jnt-SC with $\mathcal{L}_{\text{Human}}$ & Sep-SC with $\mathcal{L}_{\text{Dynamic}}$ & Jnt-SC with $\mathcal{L}_{\text{Dynamic}}$ \\

        \begin{overpic}[width=0.19\linewidth]{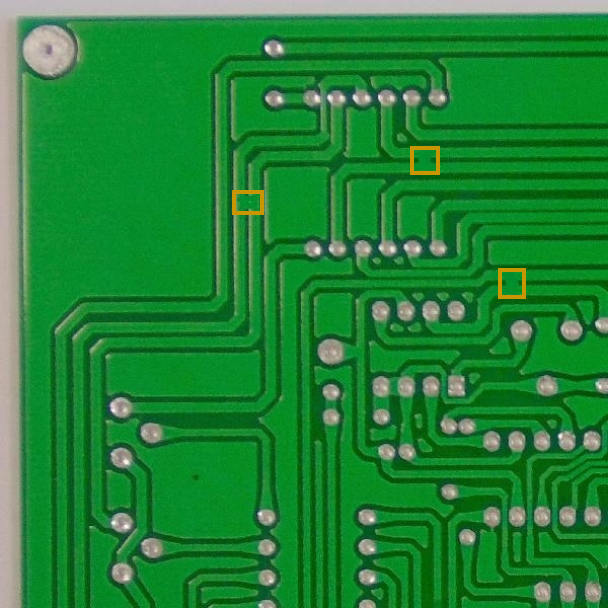}
            \put(35,72){\colorbox[RGB]{255,140,0}{\scriptsize\textcolor{white}{spur}}}
            \put(65,79){\colorbox[RGB]{255,140,0}{\scriptsize\textcolor{white}{spur}}}
            \put(78,59){\colorbox[RGB]{255,140,0}{\scriptsize\textcolor{white}{spur}}}
        \end{overpic} &
        \begin{overpic}[width=0.19\linewidth]{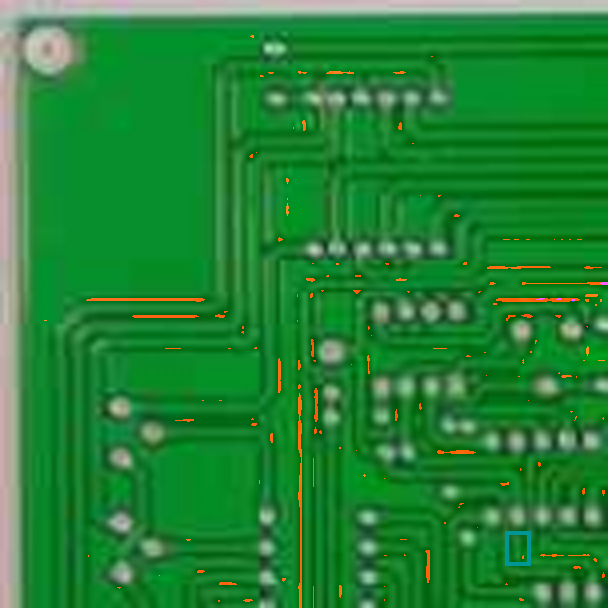}
            \put(50,15){\colorbox[RGB]{0,150,150}{\scriptsize\textcolor{white}{spurious\_copper}}}
        \end{overpic} &
        \begin{overpic}[width=0.19\linewidth]{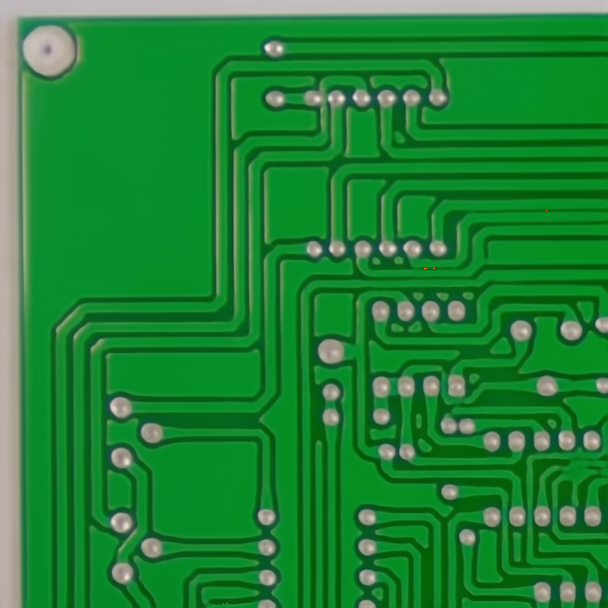}
        \end{overpic} &
        \begin{overpic}[width=0.19\linewidth]{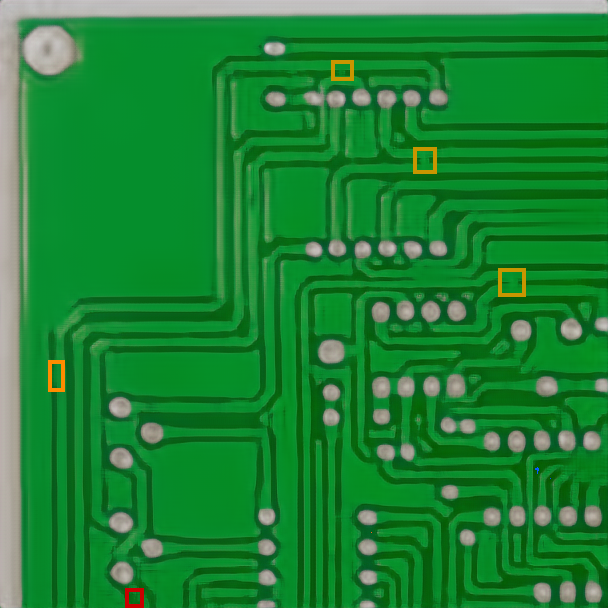}
            \put(50,90){\colorbox[RGB]{255,140,0}{\scriptsize\textcolor{white}{spur}}}
            \put(65,79){\colorbox[RGB]{255,140,0}{\scriptsize\textcolor{white}{spur}}}
            \put(78,59){\colorbox[RGB]{255,140,0}{\scriptsize\textcolor{white}{spur}}}
            \put(5,42){\colorbox[RGB]{255,140,0}{\scriptsize\textcolor{white}{spur}}}
            \put(10,6){\colorbox[RGB]{200,0,0}{\scriptsize\textcolor{white}{mouse\_bite}}}
        \end{overpic} &
        \begin{overpic}[width=0.19\linewidth
        % ,grid,tics=10,
        ]{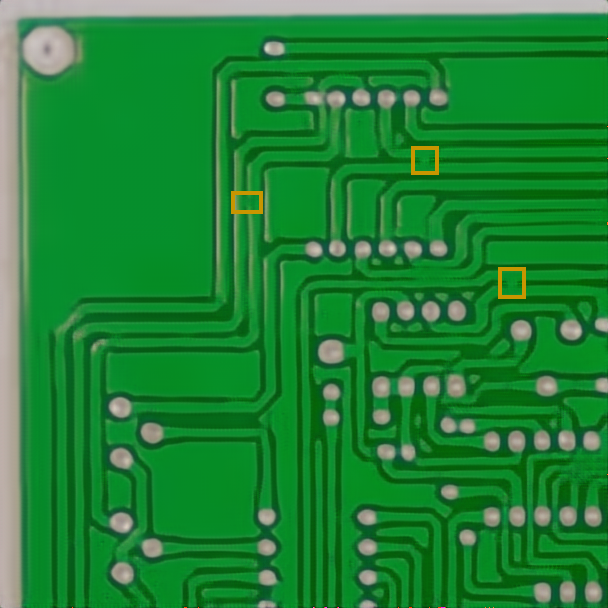}
            % \put(35, 72){\scriptsize \textcolor[RGB]{255,140,0}{spur}}
            % \put(65, 79){\scriptsize \textcolor[RGB]{255,140,0}{spur}}
            % \put(78, 59){\scriptsize \textcolor[RGB]{255,140,0}{spur}}
            \put(35,72){%
                \colorbox[RGB]{255,140,0}{\scriptsize\textcolor{white}{spur}}}
            \put(65,79){%
                \colorbox[RGB]{255,140,0}{\scriptsize\textcolor{white}{spur}}}
            \put(78,59){%
                \colorbox[RGB]{255,140,0}{\scriptsize\textcolor{white}{spur}}}
        \end{overpic} \\

        % \includegraphics[width=0.19\linewidth]{figures/4_HR.png} &
        % \includegraphics[width=0.19\linewidth]{figures/4_LR.png} &
        % \includegraphics[width=0.19\linewidth]{figures/4_JntSC_Human.png} &
        % \includegraphics[width=0.19\linewidth]{figures/4_SepSC_UW.png} &
        % \includegraphics[width=0.19\linewidth]{figures/4_JntSC_UW.png}\\
        % \multicolumn{5}{l}{\tiny The object color is:
        % \textcolor[RGB]{200,0,0}{mouse\_bite}, 
        % \textcolor[RGB]{255,140,0}{spur}, 
        % \textcolor[RGB]{0,0,200}{missing\_hole}, 
        % \textcolor[RGB]{200,150,0}{short}, 
        % \textcolor[RGB]{180,0,180}{open\_circuit},  
        % \textcolor[RGB]{0,150,150}{spurious\_copper}.
        % }
    \end{tabular}
    \caption{Qualitative comparison of detection results of different QE pipelines on the PCB Defect dataset with QF = 40.}
    \label{fig:visualize}
    \vspace{-0.5cm}
\end{figure*}

The training and validation processes are performed on PCB Defect~\cite{pcbdataset}, DeepPCB~\cite{deeppcb} and NEU-DET~\cite{neudet} datasets. To fully evaluate how training strategies affect the performance of QE model designs, we train all model configurations in Table~\ref{tab:model_variant} with three loss settings: $\mathcal{L}_{\text{Human}}$, $\mathcal{L}_{\text{Fixed}}$, and $\mathcal{L}_{\text{Dynamic}}$.We use Adam~\cite{adam} with a batch size of 2 and a cosine annealing learning rate from $1\times10^{-4}$ to $1\times10^{-7}$, restarting every $5\times10^{4}$ iterations, for a total of $3\times10^{5}$ iterations. The QE output images are evaluated using an object detection model YOLOv8~\cite{yolov8}, with mAP@50 and F1-score as the evaluation metrics for machine vision performance. 
We additionally report the number of parameters, Floating-Point-Operations (FLOPs), and inference time to analyze the computational overhead of each QE configuration.
Additionally, in all tables, the best and second-best results are highlighted in \textcolor{red}{red} and \textcolor{blue}{blue}, respectively.

% This paper examines the effectiveness of SR and CAR networks in enhancing the quality of reconstructed images for industrial applications under six distinct configurations, as outlined in Table~\ref{tab:model_variant}. \\
% \textbf{Single QE model:} The QE model simply consists of either the SR network or the CAR network alone.\\
% \textbf{Combined QE model with separate training:} The QE model is a simple combination of SR and CAR. In this configuration, SR and CAR are two separate, independently trained models. We evaluate two sequential orders: SR followed by CAR (Sep-SC), and CAR followed by SR (Sep-CS).\\
% \textbf{Combined QE model with joint training:} In contrast to the independent training of SR and CAR models, the combined QE model is trained using our proposed joint training strategy. We evaluate the same two sequential orders: SR followed by CAR (Jnt-SC), and CAR followed by SR (Jnt-CS).

\subsection{Experimental Results}

\textbf{Effect of Joint Training Strategy on Machine Vision Performance:} Table~\ref{tab:map50} reports mAP@50 and F1-scores of the downstream machine vision task (i.e., defect detection) on PCB Defect~\cite{pcbdataset}, DeepPCB~\cite{deeppcb}, and NEU-DET~\cite{neudet} datasets across different QFs and QE configurations, while Fig.~\ref{fig:avg_map} shows the corresponding average mAP@50 results over all QFs on the PCB Defect dataset. It can be seen that the combined QE models with our joint training loss (i.e., Jnt-CS and Jnt-SC) generally achieve superior performance across datasets and quality factors. Notably, the performance gains become more significant as the compression level increases (i.e., lower QF values), indicating that the proposed joint optimization is particularly effective under severe compression degradation. In contrast, the separately combined models (i.e., Sep-CS and Sep-SC) exhibit limited or even degraded performance compared with the end-to-end joint models across all experimental settings. 
\begin{table}[!t]
\centering
\caption{PSNR comparison on the PCB Defect dataset across different QE configurations.}
\label{tab:psnr}
\resizebox{\linewidth}{!}{%

\begin{tabular}{clcccccc} 
\toprule
QF                  & Loss Type                                              & SR only                                          & CAR only      & Sep-CS                  & Sep-SC                  & \makecell[c]{Jnt-CS}                       & \makecell{Jnt-SC}                                  \\
\midrule
\multirow{3}{*}{20} & $\mathcal{L}_\text{Human}$   & \textcolor{red}{29.15}  & \textcolor{red}{26.76}  & \textcolor{red}{26.99}  & \textcolor{red}{26.84}  & \textcolor{red}{32.24}      & \textcolor{red}{32.31}   \\
                    & $\mathcal{L}_\text{Fixed}$   & 26.36                   & 25.93                   & 25.77                   & 25.37                   & 26.63                       & 26.78                    \\
                    & $\mathcal{L}_\text{Dynamic}$ & \textcolor{blue}{27.18} & \textcolor{blue}{26.24} & \textcolor{blue}{26.27} & \textcolor{blue}{25.96} & \textcolor{blue}{27.77}     & \textcolor{blue}{27.93}  \\ 
\midrule
\multirow{3}{*}{40} & $\mathcal{L}_\text{Human}$   & \textcolor{red}{31.21}  & \textcolor{red}{27.43}  & \textcolor{red}{27.99}  & \textcolor{red}{28.33}  & \textcolor{red}{32.30}      & \textcolor{red}{32.54}   \\
                    & $\mathcal{L}_\text{Fixed}$   & 28.24                   & 27.00                   & 27.15                   & 27.19                   & 28.39                       & 28.79                    \\
                    & $\mathcal{L}_\text{Dynamic}$ & \textcolor{blue}{28.91} & \textcolor{blue}{27.15} & \textcolor{blue}{27.36} & \textcolor{blue}{27.75} & \textcolor{blue}{29.23}     & \textcolor{blue}{29.61}  \\ 
\midrule
\multirow{3}{*}{60} & $\mathcal{L}_\text{Human}$   & \textcolor{red}{31.88}  & \textcolor{red}{27.91}  & \textcolor{red}{28.24}  & \textcolor{red}{28.90}  & \textcolor{red}{32.36}      & \textcolor{red}{33.72}   \\
                    & $\mathcal{L}_\text{Fixed}$   & 29.16                   & 27.47                   & 27.76                   & 28.14                   & 29.24                       & 29.62                    \\
                    & $\mathcal{L}_\text{Dynamic}$ & \textcolor{blue}{29.78} & \textcolor{blue}{27.56} & \textcolor{blue}{27.90} & \textcolor{blue}{28.58} & \textcolor{blue}{30.04}     & \textcolor{blue}{30.57}  \\ 
\midrule
\multirow{3}{*}{80} & $\mathcal{L}_\text{Human}$   & \textcolor{red}{32.13}  & \textcolor{red}{28.20}  & \textcolor{red}{28.28}  & \textcolor{red}{29.75}  & \textcolor{red}{33.83}      & \textcolor{red}{34.11}   \\
                    & $\mathcal{L}_\text{Fixed}$   & 30.28                   & 27.96                   & 28.24                   & 29.32                   & 30.18                       & 30.68                    \\
                    & $\mathcal{L}_\text{Dynamic}$ & \textcolor{blue}{30.84} & \textcolor{blue}{27.99} & \textcolor{blue}{28.16} & \textcolor{blue}{29.52} & \textcolor{blue}{31.13}     & \textcolor{blue}{31.54}  \\
\bottomrule
\end{tabular}%
}
% \scriptsize 
% \begin{flushleft}
% The \textcolor{red}{Red} and \textcolor{blue}{Blue} denote the best and second best results.
% \end{flushleft}
% \vspace{-0.4cm}
\end{table}
% \begin{figure}[!t]
%     \centering
%     \includegraphics[width=1\linewidth]{figures/ablation_fixed_dynamic.png}
%     \caption{ \text{\color{red}{Model}} methods.......}
%     % \vspace{-1em}
%     \label{fig:ablation_fixed}
% % \vspace{-0.5cm}
% \end{figure}

\begin{figure}[!t]
    \centering
    \begin{overpic}[width=1\linewidth
    % ,grid,tics=10,
    ]{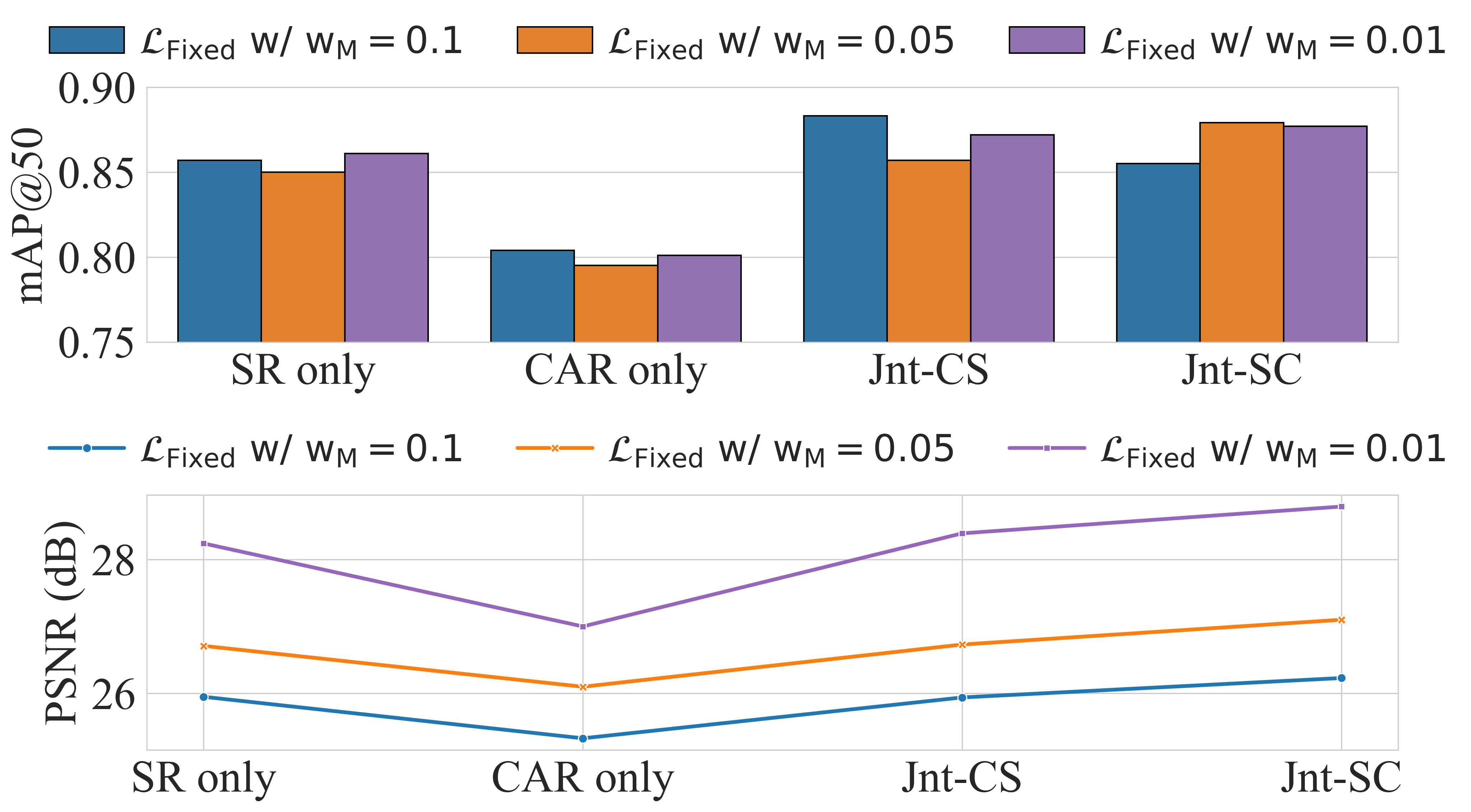}
        % ==========================
        %  Chèn text lên ảnh tại tọa độ % (x,y)
        % ==========================
        % \put(9, 53.5){\scriptsize $\mathcal{L}_\text{Fixed}$ w/ $w_M=0.1$}
        % \put(42, 53.5){\scriptsize $\mathcal{L}_\text{Fixed}$ w/ $w_M=0.05$}
        % \put(75.5, 53.5){\scriptsize $\mathcal{L}_\text{Fixed}$ w/ $w_M=0.01$}
        % \put(9, 24){\scriptsize $\mathcal{L}_\text{Fixed}$ w/ $w_M=0.1$}
        % \put(42, 24){\scriptsize $\mathcal{L}_\text{Fixed}$ w/ $w_M=0.05$}
        % \put(75.5, 24){\scriptsize $\mathcal{L}_\text{Fixed}$ w/ $w_M=0.01$}
        % \put(60,60){\scriptsize $\mathcal{L}_\text{Dynamic}$}
        % \put(60,30){\color{blue}Dynamic}
        % \put(20,50){\fcolorbox{black}{yellow!30}{\textbf{Note}}}
    \end{overpic}
    \caption{Effect of the fixed machine-perception weight $w_M$ on mAP@50 and PSNR at QF = 40.}
    \label{fig:ablation_fixed}
    \vspace{-0.5cm}
\end{figure}
% \vspace{-0.2cm}
\begin{table}[!t]
\centering
% \caption{Complexity analysis of our framework.}
\caption{Computational complexity of the evaluated QE configurations and YOLOv8 detector.}
% \color{red}

\label{tab:complexity}
\resizebox{\linewidth}{!}{%
\begin{tabular}{cccccc} 
\toprule
                                            &  CAR Architecture                       & Model Type & Parameters (M) & FLOPs (G) & Time (s) \\ 
\midrule
\multirow{7}{*}{\rotatebox[origin=c]{90}{QE Models}} & N/a                                     & SR only    & 1.60       & 48.0 & 0.0153      \\ 
\cmidrule(lr){2-6}
                                            & \multirow{3}{*}{CNN-based~\cite{nafnet}}          & CAR only   & 1.26       & 0.6        & 0.0188\\
                                            &                                         & Jnt-CS     & 2.87       & 48.6       & 0.0227\\
                                            &                                         & Jnt-SC     & 2.87       & 56.0       & 0.0237\\ 
\cmidrule(lr){2-6}
                                            & \multirow{3}{*}{Transformer-based~\cite{retinexformer}} & CAR only   & 1.61       & 4.7       & 0.0213 \\
                                            &                                         & Jnt-CS     & 3.21       & 52.7      & 0.0258 \\
                                            &                                         & Jnt-SC     & 3.21       & 123.6     & 0.0905 \\ 
\midrule \midrule
\multicolumn{2}{c}{Downstream task model}                                             & YOLOv8     & 3.16       & 4.0       & 0.0205 \\
\bottomrule
\end{tabular}
}
\vspace{1.5mm}

\noindent
\begin{minipage}{\linewidth}
\scriptsize
% \fontsize{6.5pt}{7.5pt}\selectfont

% \setlength{\baselineskip}{9.5pt}
The QE and downstream task model use input sizes of
$152\times152$ and $608\times608$, respectively. Inference
time is averaged on an NVIDIA Tesla V100-SXM2-32GB GPU.
\end{minipage}

\vspace{-0.1cm}
\end{table}

% \textbf{Effect of Joint Training Strategy on Human Perception:} Table~\ref{tab:psnr} presents a quantitative comparison of PSNR across different model configurations under three loss settings in the PCB Defect dataset~\cite{pcbdataset}. Models trained solely with the human-perception loss $\mathcal{L}_{\text{Human}}$ consistently achieve higher PSNR than those using combined loss functions (i.e., $\mathcal{L}_{\text{Fixed}}$ and $\mathcal{L}_{\text{Dynamic}}$), as $\mathcal{L}_{\text{Human}}$ directly minimizes pixel-wise reconstruction error, which aligns with the PSNR objective. In contrast, Fixed and Dynamic strategies jointly optimize reconstruction quality and downstream task performance, resulting in a slight PSNR trade-off. Among the combined losses, $\mathcal{L}_{\text{Dynamic}}$ weighting consistently outperforms the $\mathcal{L}_{\text{Fixed}}$ strategy, demonstrating a more effective balance between reconstruction quality and task-driven optimization.
% For further qualitative analysis, Fig.~\ref{fig:visualize} shows defect detection results obtained with different QE pipelines on the PCB Defect dataset at QF~40. While Jnt-SC trained with $\mathcal{L}_{\text{Human}}$ yields high PSNR, it fails to perform reliable defect detection due to its human-perception-oriented optimization. Conversely, Jnt-SC model optimized with  $\mathcal{L}_{\text{Dynamic}}$ identifies all defects despite lower PSNR, highlighting machine-perception objectives are essential for industrial inspection.

\textbf{Effect of Joint Training Strategy on Human Perception:} 
Table~\ref{tab:psnr} compares PSNR across model configurations and loss settings on the PCB Defect dataset~\cite{pcbdataset}. Models trained only with $\mathcal{L}_{\text{Human}}$ achieve the highest PSNR because this objective directly minimizes pixel-wise reconstruction error. In contrast, $\mathcal{L}_{\text{Fixed}}$ and $\mathcal{L}_{\text{Dynamic}}$ balance reconstruction quality with downstream performance, resulting in lower PSNR. Among them, $\mathcal{L}_{\text{Dynamic}}$ consistently achieves a better balance. Fig.~\ref{fig:visualize} further compares detection results at QF = 40. Although Jnt-SC trained with $\mathcal{L}_{\text{Human}}$ produces higher PSNR, it misses defects. Conversely, $\mathcal{L}_{\text{Dynamic}}$ detects all defects despite lower PSNR, confirming the importance of machine-perception supervision for industrial inspection.
Beyond perceptual and detection performance, Table~\ref{tab:complexity} shows that the additional QE overhead yields substantial accuracy gains, increasing mAP@50 from 0.436 without QE to 0.965 with Jnt-SC at QF = 80. Nevertheless, reducing this overhead while preserving detection performance remains an important direction for future work.
% }

\subsection{Ablation Studies}

\textbf{Fixed-Weight Loss Function with Varying Weight Values:} 
Fig.~\ref{fig:ablation_fixed} evaluates machine-perception weights $w_M$ of 0.1, 0.05, and 0.01 on the PCB Defect dataset at QF = 40. Although $w_M=0.01$ slightly reduces mAP@50 in some configurations, the gap is small, while lower $w_M$ consistently improves PSNR, indicating a detection--fidelity trade-off. Since $w_M=0.01$ provides clear PSNR gains with minor mAP@50 loss, we adopt this setting in all experiments involving $\mathcal{L}_{\text{Fixed}}$.
% \textbf{Analysis of Machine Perception Loss Components:} The machine-perception loss $\mathcal{L}_{\text{Machine}}$ consists of classification loss $\mathcal{L}_{\text{cls}}$, localization losses $\mathcal{L}_{\text{box}}$ and $\mathcal{L}_{\text{dfl}}$, where $\mathcal{L}_{\text{dfl}}$ enhances bounding box regression by modeling localization as a discrete distribution. 
% As shown in Table~\ref{tab:ablation_loss_type}, jointly optimizing all loss components yields the best performance compared to partial loss configurations. Using only $\mathcal{L}_{\text{cls}}$ outperforms the localization-only objective but remains inferior to the full $\mathcal{L}_{\text{Machine}}$. These results confirm the necessity of localization and classification supervision for detection performance.

% \usepackage{color}
% \usepackage{graphicx}
% \usepackage{multirow}

\begin{table}[!t]
\centering
% \caption{Analysis of Machine Loss components impact on QE performance at QF = 40.}
\caption{Effect of machine-perception loss components on mAP@50 at QF = 40.}
\label{tab:ablation_loss_type}
\resizebox{\linewidth}{!}{%
\begin{tabular}{lcccccccc} 
\toprule
\multirow{2}{*}{Loss Type}       & \multicolumn{2}{c}{Loss Element}      & \multirow{2}{*}{SR only}                    &\multirow{2}{*}{CAR only}                  & \multirow{2}{*}{Sep-CS} & \multirow{2}{*}{Sep-SC} & \multirow{2}{*}{Jnt-CS}                & \multirow{2}{*}{Jnt-SC}                 \\
                                 & $\mathcal{L}_{\text{box}}+\mathcal{L}_{\text{dfl}}$ & $\mathcal{L}_{\text{cls}}$ &                   &                  &                         &                         &                    &                     \\ 
\midrule
$\mathcal{L}_\text{Human}$ &                   &                   & 0.476                   & 0.330                   & 0.318                   & 0.461                   & 0.395                   & 0.440                    \\ 
\midrule
\multirow{3}{*}{$\mathcal{L}_\text{Fixed}$}           & \checkmark        &                   & 0.763                   & 0.665                   & 0.617                   & 0.704                   & 0.772                   & 0.804                    \\
                                 &                   & \checkmark        & \textcolor{blue}{0.852} & \textcolor{blue}{0.780} & \textcolor{blue}{0.721} & \textcolor{blue}{0.787} & \textcolor{blue}{0.862} & \textcolor{blue}{0.876}  \\
                                 & \checkmark        & \checkmark        & \textcolor{red}{0.861}  & \textcolor{red}{0.801}  & \textcolor{red}{0.710}  & \textcolor{red}{0.764}  & \textcolor{red}{0.872}  & \textcolor{red}{0.877}   \\ 
\midrule
                                 & \checkmark        &                   & 0.765                   & 0.676                   & 0.624                   & 0.706                   & 0.767                   & 0.795                    \\
$\mathcal{L}_\text{Dynamic}$                          &                   & \checkmark        & \textcolor{blue}{0.848} & \textcolor{blue}{0.774} & \textcolor{blue}{0.677} & \textcolor{blue}{0.777} & \textcolor{blue}{0.866} & \textcolor{blue}{0.866}  \\
                                 & \checkmark        & \checkmark        & \textcolor{red}{0.866}  & \textcolor{red}{0.804}  & \textcolor{red}{0.741}  & \textcolor{red}{0.814}  & \textcolor{red}{0.876}  & \textcolor{red}{0.887}   \\
\bottomrule
\end{tabular}%
}
% \scriptsize 
% \begin{flushleft}
% The \textcolor{red}{Red} and \textcolor{blue}{Blue} denote the best and second best results.
% \end{flushleft}
\vspace{-0.2cm}
\end{table}
\textbf{Analysis of Machine Perception Loss Components:} 
The machine-perception loss $\mathcal{L}_{\text{Machine}}$ combines classification loss $\mathcal{L}_{\text{cls}}$ and localization losses $\mathcal{L}_{\text{box}}$ and $\mathcal{L}_{\text{dfl}}$, with $\mathcal{L}_{\text{dfl}}$ modeling bounding-box localization as a discrete distribution. Table~\ref{tab:ablation_loss_type} shows that jointly optimizing all components performs best. Although $\mathcal{L}_{\text{cls}}$ alone outperforms the localization-only objective, it remains inferior to the full objective, confirming the necessity of both classification and localization supervision.

% \subsubsection{Model architecture for CAR model}
% \input{chapters/table/ablation_qe_type}
% Table~\ref{tab:ablation_qe_type}

% \textbf{Performance -- Complexity Trade-off:} Table~\ref{tab:complexity} presents the trade-off between model complexity and downstream detection performance on the PCB Defect dataset~\cite{pcbdataset} at QF value of 40. 
% Among all configurations, the transformer-based CAR~\cite{retinexformer} with the Jnt-SC setting achieves the highest mAP@50 of 0.887. 
% However, this performance gain comes at a substantially higher computational cost, requiring 123.6~GFLOPs, which is more than twice that of the CNN-based CAR~\cite{nafnet} model achieving a comparable mAP@50 of 0.877 with only 56.0~GFLOPs. 
% These results highlight that CNN-based CAR~\cite{nafnet} models offer a more favorable efficiency-accuracy balance, while transformer-based approaches~\cite{retinexformer} generally emphasize higher peak performance with increased computational cost.
% \textbf{Performance Analysis of CAR Architectures:}
% Table~\ref{tab:car_architecture} shows the performance comparison of QE models trained with $\mathcal{L}_{\text{Dynamic}}$ across different CAR architectures. As illustrated, joint training strategies (Jnt-CS and Jnt-SC) consistently achieve the best mAP@50 performance in both CNN-based and Transformer-based architectures, surpassing separate and standalone models. These results confirm the effectiveness of the proposed joint training among various CAR architectures.

\textbf{Ablation with Different CAR Models:}
Table~\ref{tab:car_architecture} compares the performance of QE frameworks trained with $\mathcal{L}_{\text{Dynamic}}$ across different CAR architectures. As shown, applying our joint training strategy consistently achieves the best mAP@50 performance for both CNN-based and transformer-based CAR models, outperforming separate and standalone configurations. These results demonstrate that the proposed joint training strategy generalizes effective across both evaluated architectures

% \textbf{Effect of the detection model:} Table \ref{tab:detection} presents the detection performance of the Jnt-SC configuration at QF = 40 on the PCB Defect dataset. In this experiment, the Jnt-SC model is trained with Uncertainty Weighting (UW), where the machine loss is generated solely from the YOLOv8 detector. The other detectors, YOLOv11 and Faster R-CNN, are pretrained on the original PCB Defect dataset and are used only for evaluation. As shown in Table \ref{tab:detection}, the Jnt-SC model trained with YOLOv8 still yields substantial performance gains when evaluated using different detectors such as YOLOv11 and Faster R-CNN. This demonstrates that our restoration network is not tightly coupled to a specific detector. Instead, the task-oriented optimization encourages the QE model to enhance semantic information that is generally beneficial for object detection, rather than learning detector-specific patterns.
% Therefore, even when the detector changes, the restoration network does not necessarily need to be retrained, and the improvements generalize well across different detection models. We will clarify this point in the revised manuscript.

% \usepackage{color}
% \usepackage{graphicx}
% \usepackage{multirow}

\begin{table}[!t]
\centering
% \caption{mAP@50 comparison across different CAR architectures optimized with $\mathcal{L}_{\text{Dynamic}}$ at QF = 40.}
\caption{mAP@50 comparison of different CAR architectures using $\mathcal{L}_{\text{Dynamic}}$ at QF = 40.}
\label{tab:car_architecture}
\resizebox{\linewidth}{!}{%

\begin{tabular}{lcccccc} 
\toprule
CAR Architecture                              & SR only & CAR only & Sep-CS & Sep-SC & \makecell[c]{Jnt-CS}    & \makecell{Jnt-SC}        \\ 
\midrule
CNN-based~\cite{nafnet}                       & \multirow{2}{*}{0.866}   & 0.668    & 0.699  & 0.811  & \textcolor{red}{0.877}  & \textcolor{blue}{0.871}  \\ 
Transformer-based~\cite{retinexformer} &                          & 0.804    & 0.741  & 0.814  & \textcolor{blue}{0.876} & \textcolor{red}{0.887}   \\
\bottomrule
\end{tabular}%
}
% \scriptsize 
% \begin{flushleft}
% The \textcolor{red}{Red} and \textcolor{blue}{Blue} denote the best and second best results.
% \end{flushleft}
% \vspace{-0.2cm}
\end{table}
% \vspace{-0.2cm}
\begin{table}[!t]
\centering
% \scriptsize
% \color{red}

\caption{Detection performance with various types of detection models at QF = 40 on the PCB Defect dataset.}
% \caption{mAP@50 of Jnt-SC evaluated with different detectors on the PCB Defect dataset at QF = 40.}
\label{tab:detection}
% \resizebox{\linewidth}{!}{%
% \renewcommand{\arraystretch}{0.8}
\begin{tabular}{lccc}
\toprule
       & YOLOv8 (Ours) & YOLOv11 & Faster R-CNN \\ \midrule
Jnt-SC w/o UW & 0.440   & 0.457   & 0.310      \\
Jnt-SC w/ UW  & \textbf{0.887}   & \textbf{0.871}   & \textbf{0.693}      \\ \bottomrule
\end{tabular}%
% }
\vspace{-0.2cm}
\end{table}
% \textcolor{red}{
% \textbf{Effect of the detection model:} Table \ref{tab:detection} reports the detection performance of Jnt-SC at QF = 40 on the PCB Defect dataset. Jnt-SC is trained using Uncertainty Weighting (UW), with the machine loss derived only from YOLOv8, while pretrained YOLOv11 and Faster R-CNN models are used solely for evaluation. Despite being optimized with YOLOv8, Jnt-SC consistently improves detection performance across all evaluated detectors. This indicates that the restoration network is not detector-specific; instead, task-oriented optimization enhances semantic information that is broadly useful for object detection. Therefore, changing the detector does not necessarily require retraining the restoration network. 
\textbf{Ablation with Different Detection Models:} Table~\ref{tab:detection} reports Jnt-SC performance at QF = 40 on the PCB Defect dataset. Jnt-SC is trained with UW using machine loss from YOLOv8, while YOLOv11 and Faster R-CNN are used only for evaluation. Nevertheless, it improves detection across all detectors, showing that task-oriented optimization enhances features that transfer across different detector architectures. Therefore, changing the detector may not require retraining the restoration network.

\section{Conclusions}
\label{sec:conclusions}
In this paper, we propose a novel quality enhancement framework that aims to reconstruct high-resolution images with minimal compression artifacts from compressed low-resolution inputs, specifically for downstream machine vision tasks. 
Our approach introduces an end-to-end joint model that integrates SR and CAR through multiple architectural variants. 
Moreover, we design a unified training strategy with an integrated loss function that jointly optimizes human-perception and machine-perception objectives. 
Our extensive experiments illustrate the effectiveness and efficiency of our method across various datasets and configurations. 
\section*{Acknowledgement}
This research is funded by Vietnam National Foundation for Science and Technology Development (NAFOSTED) under grant number NCUD.02-2024.09.
% \newpage
\bibliographystyle{IEEEtran}
\bibliography{IEEEabrv,icme_paper}

\end{document}